\documentclass[10pt,twocolumn,letterpaper]{article}

\usepackage{iccv}
\usepackage{times}
\usepackage{epsfig}
\usepackage{graphicx}
\usepackage{amsmath}
\usepackage{amssymb}

\usepackage[nospace]{cite}
\usepackage[tight,footnotesize]{subfigure}
\usepackage{multirow}

\usepackage[pagebackref=true,breaklinks=true,letterpaper=true,colorlinks,bookmarks=false]{hyperref}

\iccvfinalcopy 

\def\iccvPaperID{****} 

\ificcvfinal\fi
\begin{document}

\title{GraphX-Convolution for Point Cloud Deformation in 2D-to-3D Conversion}

\author{Anh-Duc Nguyen
\qquad
Seonghwa Choi
\qquad
Woojae Kim
\qquad
Sanghoon Lee\\
Yonsei University\\
{\tt\small \{adnguyen,csh0772,wooyoa,slee\}@yonsei.ac.kr}
}

\maketitle

\begin{abstract}
In this paper, we present a novel deep method to reconstruct a point cloud of an object from a single still image. Prior arts in the field struggle to reconstruct an accurate and scalable 3D model due to either the inefficient and expensive 3D representations, the dependency between the output and number of model parameters or the lack of a suitable computing operation.
We propose to overcome these by deforming a random point cloud to the object shape through two steps: feature blending and deformation. In the first step, the global and point-specific shape features extracted from a 2D object image are blended with the encoded feature of a randomly generated point cloud, and then this mixture is sent to the deformation step to produce the final representative point set of the object. 
In the deformation process, we introduce a new layer termed as GraphX that considers the inter-relationship between points like common graph convolutions but operates on unordered sets. Moreover, with a simple trick, the proposed model can generate an arbitrary-sized point cloud, which is the first deep method to do so. Extensive experiments verify that we outperform existing models and halve the state-of-the-art distance score in single image 3D reconstruction.
\end{abstract}

\section{Introduction}
\label{intro}

Our world is 3D, and so is our perception. Making machines see the world like us is the ultimate goal of computer vision. So far, we have made significant advancements in 2D machine vision tasks, and yet 3D reasoning from 2D still remains very challenging. 3D shape reasoning is of the utmost importance in computer vision as it plays a vital role in robotics, modeling, graphics, and so on. Currently, given multiple images from different viewpoints, computers are able to estimate a reliable shape of the interested object. Yet, when we humans look at a single 2D image, we still understand the underlying 3D space to some extent thanks to our experience, but machines are nowhere near our perception level. Thus, a crucial yet demanding question is whether we can help machines to achieve a similar 3D understanding and reasoning ability of humans. 
At first, a solution seems highly unlikely because some information is permanently lost as we go from 3D to 2D. However, if a machine is able to learn a shape prior like humans, then it can infer 3D shape from 2D effortlessly.

\begin{figure}
\centering
\includegraphics[width=0.5\textwidth]{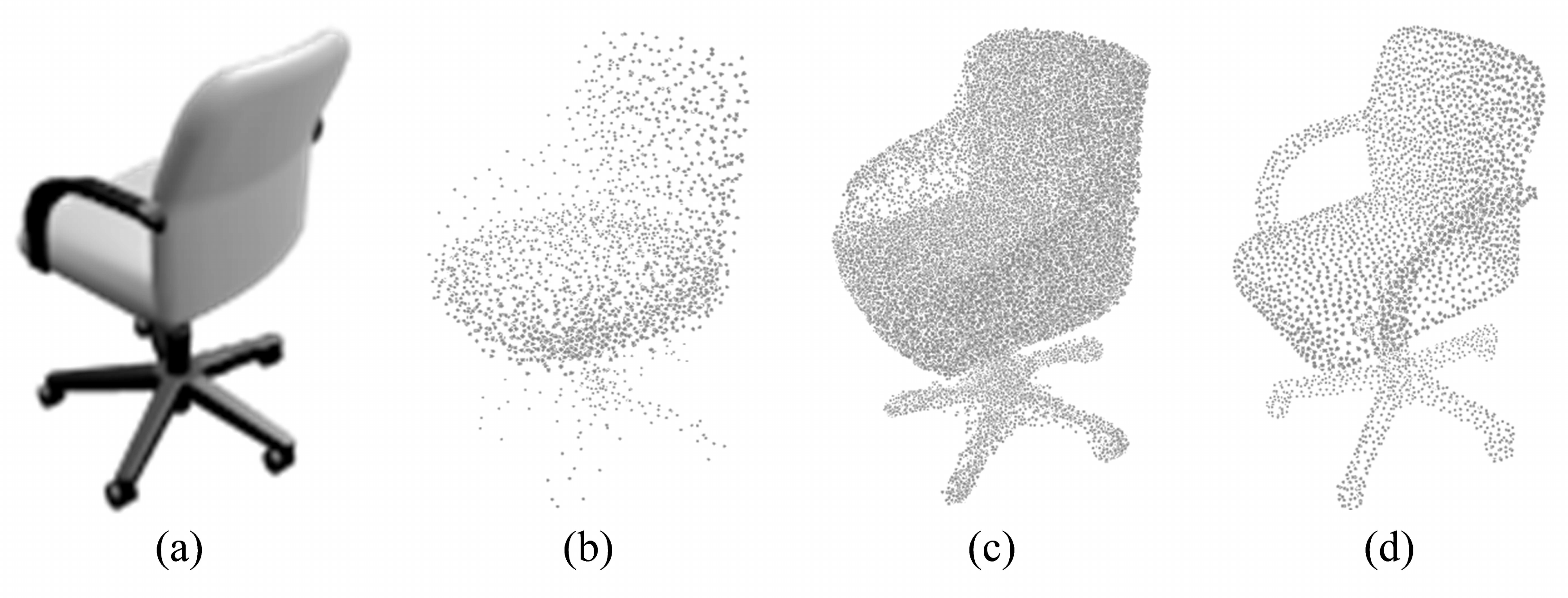}
\caption{A sample showing what our model is capable of. (a) An RGB input image. (b) A 2k-point. (c) A 40k-point clouds produced by PCDNet. (d) A ground truth point cloud of the model.}
\label{intro:sample}
\end{figure}

Deep learning, or most of the time, deep convolutional neural networks (CNNs), has recently shown a promising learning ability in computer vision. However, there is not yet an easy and efficient way to apply deep learning to 3D reconstruction. Most modern progress of deep learning is in areas where signals are ordered and regular -- images, audios and languages to name a few, while common 3D representations such as meshes or point clouds are unordered and irregular. Therefore, there is no guarantee that all the bells and whistles from the 2D practice would work in the 3D counterpart. Other 3D structures may result in easier learning such as grid voxels but at the cost of computational efficiency. Also, quantization errors in these structures may diminish the natural invariances of the data \cite{rn270}.

In this regard, we present a novel deep method to reconstruct a 3D point cloud representation of an object from a single 2D image.
Even though a point cloud representation does not possess appealing 3D geometrical properties like a mesh or CAD model, it is simple and efficient when it comes to transformation and deformation, and can produce high-quality shape models \cite{rn262}. 

Our insight into prior arts leads to the realization of several key properties of the prospective system: (1) the model should make predictions based on not only local features but also high-level semantics, (2) the model should consider spatial correlation between points, and (3) the method should be scalable, \ie , the output point cloud can be of arbitrary size.
To inherit all these properties, we propose to approach the problem in two steps: \textit{feature blending} and \textit{deformation}. 
In the first step, we extract point-specific and global shape features from the 2D input object image and blend into the encoded feature of a randomly generated point cloud. 
The per-point features are obtained by a simple projection of the point cloud onto the shape features extracted from the encoded image.
For the global information, we borrow an idea from image style transfer literature that is conceptually simple and suited to our problem formulation. The per-point and global features are processed by a deformation network to produce a point cloud for the given object. 
Despite the simplicity of the global shape feature, its mere introduction already helps the proposed system to outperform the state of the art. 

To further improve on this baseline, in the deformation step, we introduce a new layer termed as GraphX that learns the inter-relationship among points like common graph convolutions \cite{rn261} but can operate on unordered point sets. GraphX also linearly combines points similar to $\mathcal{X}$-convolution \cite{rn260} but in a more global scale. 
Armed with more firepower, our model surpasses all the existing single image 3D reconstruction methods and reduces the current state-of-the-art distance metric to half.
Finally, we showcase that the proposed model can generate an arbitrary-sized point cloud for a given object, which is the first deep method to do so according to our knowledge. 
An example of the predicted point clouds of a CAD chair model that our model produces is shown in Figure~\ref{intro:sample}.
We dub the proposed method Point Cloud Deformation NETwork (PCDNet) for brevity.

Our contributions are three-fold. 
First, we introduce a new 3D reconstruction model which is the first to generate a point cloud representation of arbitrary size. 
Second, we present a new global shape feature, which is inspired by image style transfer literature. The extraction operation is a symmetric mapping, so the network is invariant to the orderlessness of point cloud. 
Finally, we propose a new layer termed as GraphX which learns the inter-connection among points in an unordered set.
%
%
To facilitate future research, the code has been released at \url{https://github.com/justanhduc/graphx-conv}.

\begin{figure*}[h!]
\centering
\includegraphics[width=\textwidth]{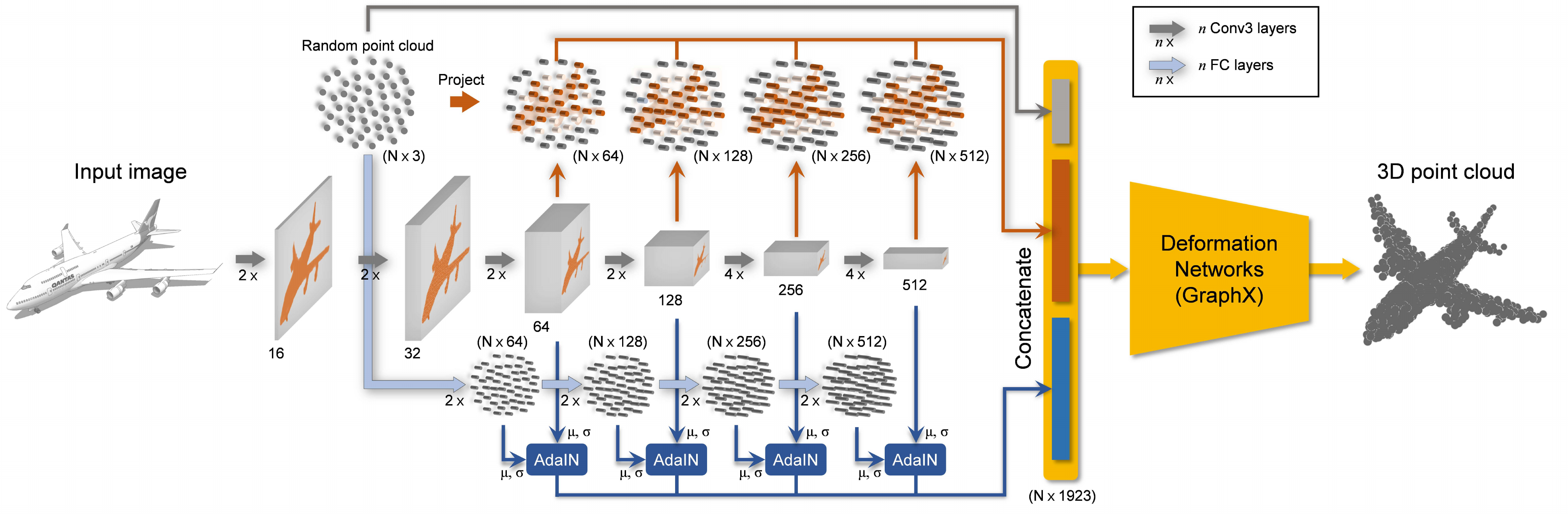}
\caption{Overview of PCDNet. The network consists of three separate branches. Image encoding: this branch (middle) is a CNN that takes an input image and encodes it into multi-scale 2D feature maps. Point-specific shape information extraction: this branch (top), which is parameter-free, simply projects the initial point set to the 2D feature maps at every scale to form point-specific features. Global shape information extraction: the final branch (bottom) is an MLP that processes a randomly generated point cloud and 2D output features from the CNN. The features and the 2D feature maps at the same scales are fed to an AdaIN operator to produce global shape features. All the features plus the point cloud are concatenated and input to a deformation network. }
\label{method:framework}
\end{figure*}

\section{Related work}
\label{rel-work}
3D reconstruction is one of the holy grail problems in computer vision.
The most traditional approach to this problem is perhaps Structure-from-Motion \cite{rn264} or Shape-from-X \cite{rn265,rn273}. However, while the former requires multiple images of the same scene from slightly different viewpoints and an excellent image matching algorithm, the latter requires prior knowledge of the light sources as well as albedo maps, which makes it suitable mainly for a studio environment. 
Some early studies also consider learning shape priors from data. Notably, Saxena et al. \cite{rn266} constructed a Markov random field to model the relationship between image depth and various visual cues to recreate a 3D ``feeling" of the scene. In a similar study, the authors in \cite{rn274} learned different semantic likelihoods to achieve the same goal.

Recently, deep learning, or most likely deep CNNs, has rapidly improved various fields \cite{Kim_Lee_2017,
Kim_Nguyen_Ahn_Luo_Lee_2018,
Kim_Nguyen_Lee_2018,
Kim_Zeng_Ghadiyaram_Lee_Zhang_Bovik_2017,
Kim_Kim_Ahn_Kim_Lee_2018,
Nguyen_Choi_Kim_Lee_2019,
rn153,
rn218,
rn92} including 3D reconstruction. 
Deep learning-based methods can reconstruct an object from a single image by learning the geometries of the object available in the image(s) and hallucinating the rest thanks to the phenomenal ability to estimate statistics from images. 
The obtained results are usually far more impressive than the traditional single image 3D reconstruction methods. 
Wu et al. \cite{rn267} employed a conditional deep belief network to model volumetric 3D shapes. Yan et al. \cite{rn268} introduced an encoder-decoder network regularized by a perspective loss to predict 3D volumetric shapes from 2D images. In \cite{rn269}, the authors utilized a generative model to generate 3D voxel objects arbitrarily. Tulsianni et al. \cite{rn275} introduced ray-tracing into the picture to predict multiple semantics from an image including a 3D voxel model. Howbeit, voxel representation is known to be inefficient and computationally unfriendly \cite{rn259,rn262}. 
For mesh representation, Wang et al. \cite{rn259} gradually deformed an elliptical mesh given an input image by using graph convolution, but mesh representation requires overhead construction, and graph convolution may result in computing redundancy as masking is needed. 
There has been a number of studies trying to reconstruct objects without 3D supervision \cite{rn263,rn276,rn277}. These methods leveraged the multi-view projections of the models to bypass the need for 3D supervising signals. The closest work to ours is perhaps Fan et al. \cite{rn262}. The authors proposed an encoder-decoder architecture with various shortcuts to directly map an input image to its point cloud representation. A disadvantage of the existing methods that directly generate point sets is that the number of trainable parameters are proportional to the number of points in the output cloud. Hence, there is always an upper bound for the point cloud size. 
In contrast, the proposed PCDNet overcomes this problem by deforming a point cloud instead of making one, which makes the system far more scalable.

\section{Point cloud deformation network}
\label{method}

Our overall framework is shown in Figure~\ref{method:framework}. Given an input object image, we first encode it by using a CNN to extract multi-scale feature maps. From these features, we further distill global and point-specific shape information of the object. 
The obtained information is then blended into a randomly generated point cloud, and the mixture is fed to a deformation network. 
All the modules are differentiable, ergo it can be trained end-to-end in any contemporary deep learning library. 
In the following sections, we will describe all the steps in details.

\subsection{Image encoding}
We use a VGG-like architecture \cite{rn64} similar to \cite{rn259} to encode the input image (Figure~\ref{method:framework} middle branch). The noteworthy aspect of the architecture is that it is a feed-forward network without any shortcut from lower layers, and it consists of several spatial downsamplings and channel upsamplings at the same time. This sort of architecture allows a multi-scale representation of the original image and has been shown to work better than the modern designs with skip connections when it comes to shape or texture representation \cite{rn259,rn237}. 

\subsection{Feature blending}
\label{method:feature}

\subsubsection{Point-specific shape information}
Following \cite{rn259}, we extract a feature vector for each individual point by projecting the points onto the feature maps as illustrated in Figure~\ref{method:framework} (top branch). Concretely, given an initial point cloud, we compute the 2D pixel coordinate of each point using camera intrinsics. Since the resulting coordinates are floating point, we resample the feature vectors using bilinear interpolation. Note that we reuse the same image feature maps for both the projection and global shape features.

\subsubsection{Global shape information}
\label{method:feature:global}
The global shape information is obtained by the bottom branch in Figure~\ref{method:framework}.
To derive the global shape information, we borrow a concept from image style transfer literature. Image style transfer concerns how a machine can artistically replicate the ``style" of an image, possibly color, textures, pen strokes, \textit{etc}., on a target image without overwriting its contents. We find an analogy between this style transfer and our problem formulation in the sense that given an initial point cloud, which is analogous to the target image in style transfer, we would like to transfer the ``style" of the object, which is the shape of the input object in our case, to the initial point set. To this end, we propose to ``stylize" the initial point cloud by the adaptive instance normalization (AdaIN)~\cite{rn218}. First, we process the initial point cloud by a simple multi-layer perceptron (MLP) encoder composed of several blocks of fully connected (FC) layers to obtain features at multiple scales. 
We note that the number of scales here is equal to that of the image feature maps, and the dimensionality of the feature is the same as the number of the feature map channels at the same scale. Let the set of $c_i$-dimensional features from the MLP and the 2D feature maps from the CNN at scale $i$ be $\mathcal{Y}_i \subseteq\mathcal{R}^{c_i}$ and $X_i\in \mathcal{R}^{c_i\times h_i\times w_i}$ ($c_i$ channels, height $h_i$, and width $w_i$), respectively. We define the 2D-to-3D AdaIN as 

\begin{equation}
\text{AdaIN}(X_i, y_j) = \sigma_{X_i}\frac{y_j-\mu_{\mathcal{Y}_i}}{\sigma_{\mathcal{Y}_i}} + \mu_{X_i},
\end{equation}
where $y_j\in \mathcal{Y}_i$ is the feature vector of point $j$ in the cloud, $\mu_{X_i}$ and $\sigma_{X_i}$ are the mean and standard deviation of $X_i$ taken over all the spatial locations, and $\mu_{\mathcal{Y}_i}$ and $\sigma_{\mathcal{Y}_i}$ are the mean and standard deviation of the point cloud in the feature space. The rationale of our definition is that from a global point of view, an object shape can be characterized by a mean shape and an associated variance. We can retrieve these mean shape and variance from the 2D input image, and then embed them into the initial 3D point cloud after ``neutralizing" it by removing its mean and variance. In Section~\ref{exp:ablation}, we will show an experiment that reinforces our view.

\subsubsection{Point cloud feature extraction}
After extracting the global and per-point features, to obtain a single feature vector for each point, we simply concatenate the two features together with the point coordinates. We note that our feature extraction is somewhat similar to that of PointNet \cite{rn270} in the sense that both methods consider global and per-point features as well as the symmetry property of the global one. Like semantic segmentation in \cite{rn270}, point cloud generation should rely on both local geometry and global semantics. Each point's position is predicted based on not only its individual feature but also the collection of points as a whole. More importantly, since the global semantics do not change as the points are permuted, the global feature must be invariant with respect to permutation. While max pooling is adopted in \cite{rn270}, which makes sense as the method emphasizes only on critical features to predict labels, we use mean and variance here because they characterize distribution naturally. 

\subsection{Point cloud deformation}

\begin{figure}
\centering
\includegraphics[width=0.5\textwidth]{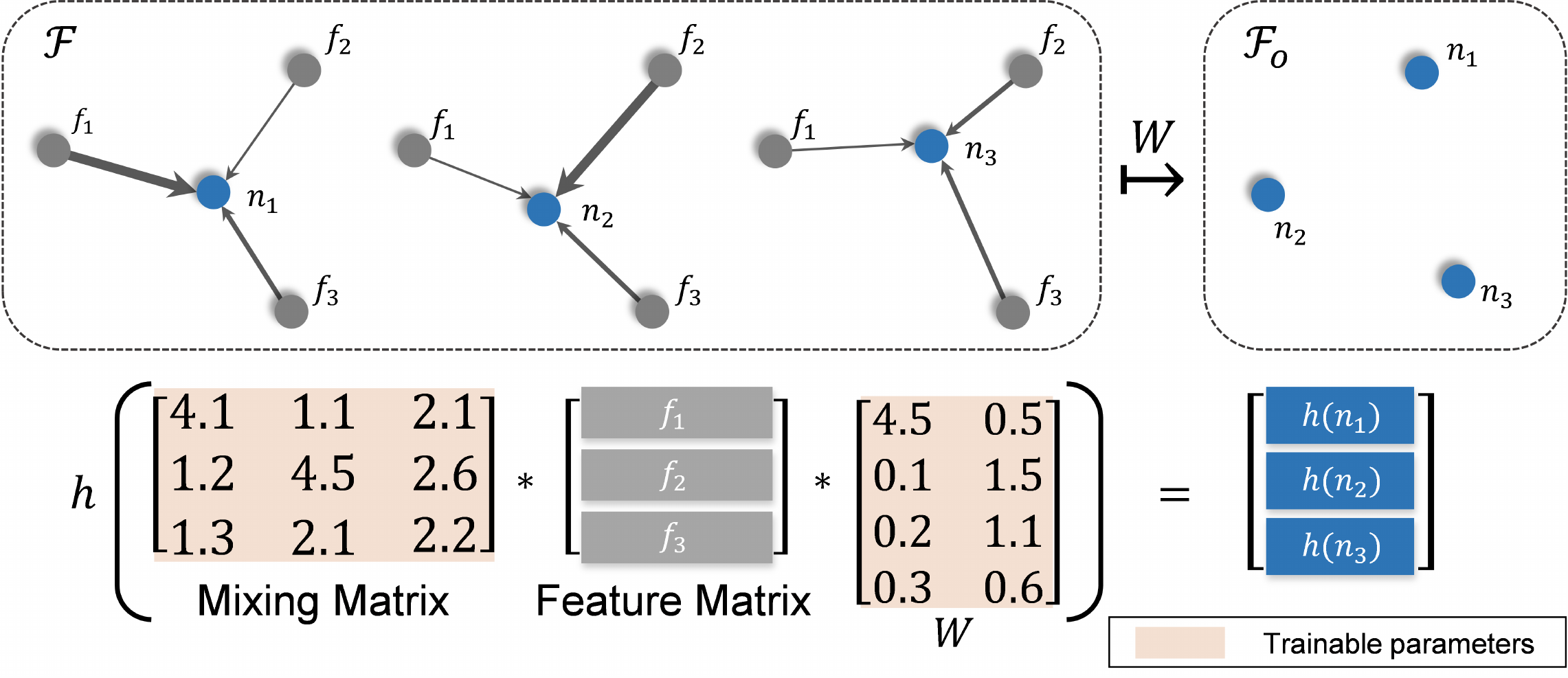}
\caption{An illustration of GraphX. First, the new points $n_k$ are computed by combining all the given points $f_i$ according to a mixing weight. Then the new points are mapped from the current space $\mathcal{F}$ to a new space $\mathcal{F}_o$ by $W$ and activated by a non-linear activation $h(\cdot)$. For brevity, biases are omitted.}
\label{mehthod:pc-deform:graphX-fig}
\end{figure}

We now proceed to the last phase of our method which produces a point cloud representation of the input object via an NN. In order to generate a precise and representative point cloud, it is necessary to establish some communication between points in the set. 
The $\mathcal{X}$-convolution ($\mathcal{X}$-conv) \cite{rn260} seemingly fits our purpose as the operator is carried out in a neighborhood of each point. However, because this operator runs a built-in K-nearest neighbor every iteration, the computational time is prohibitively long when the cloud size is large and/or the network has many $\mathcal{X}$-conv layers. On the other hand, graph convolution \cite{rn261} considers the local interaction of points (or in this case vertices) but unfortunately, the operator is designed for mesh representation which requires an adjacency matrix. Due to these shortcomings, an operator with a similar functionality but having greater freedom is required to ensure efficient learning on unordered point sets. 

In this paper, inspired by the simplicity of graph convolution and the way $\mathcal{X}$-conv works, we propose \textit{graphX-convolution} (GraphX) which possesses a similar functionality as the graph convolution but works on unordered point sets like $\mathcal{X}$-conv. An intuitive illustration of GraphX is demonstrated in Figure~\ref{mehthod:pc-deform:graphX-fig}. The operation starts by mixing the features in the input and then applies a usual FC layer. Let $\mathcal{F}_j\subseteq \mathcal{R}^{d_j}$ be the set of $d_j$-dimensional features fed to $j^{th}$ layer of the deformation network. For notation simplicity, we drop the layer index $j$ and denote the output set as $\mathcal{F}_o\subseteq \mathcal{R}^{d_o}$.  Mathematically, GraphX is defined as

\begin{equation}
\label{mehthod:pc-deform:graphX}
\begin{aligned}
f^{(o)}_k =h\left(n_k\right) = h\left( {W^T( \sum\limits_{f_i \in \mathcal{F}} {w_{ik} f_i} + b_k } ) + b \right),
\end{aligned}
\end{equation}
where $f^{(o)}_k$ is the $k^{th}$ output feature vector in $\mathcal{F}_o$, $w_{ik},b_k\in \mathcal{R}$ are trainable \textit{mixing weight} and \textit{mixing bias} corresponding to each pair $(f_i,f^{(o)}_k)$, $W\in \mathcal{R}^{d\times d_o}$ and $b\in\mathcal{R}^{d_o}$ are the weight and bias of the FC layer, and $h$ is an optional non-linear activation. The formulation of GraphX can be seen as a global graph convolution. Instead of learning weights for only neighboring points, GraphX learns for the whole point set. This definition is based on our hypothesis that in a point cloud, every point can convey more or less information about others, thus we can let the learning decide where the network should concentrate. Still, learning two full $d\times d_o$ weight matrices like the graph convolution requires mammoth computer calculations. Therefore, we break the weight into a fixed $W$ for all the points and an adaptive part, $w_{ik}$, which is just a scalar. Our method is also similar to $\mathcal{X}$-conv in the way it takes the relationship of points into account, but while the mixing matrix of $\mathcal{X}$-conv is computed by a neural network from a locality of points, ours is directly learned and works on the whole point set, and hence capable of learning a local-to-global prior. 

If the size of the point cloud is large, learning a mixing operation is still potentially expensive. One workaround is to start with a small point cloud, and then gradually upsample it in such a way that $\left|\mathcal{F}_o\right|>\left|\mathcal{F}\right|$. Thus, the computation and memory can be reduced considerably. Alternatively, GraphX can also be utilized in the downsampling direction which is useful in point cloud encoding.

To further reduce the memory footprint, we propose to decompose $W$ into a low-rank representation. For \eg, we can replace $W\in \mathcal{R}^{d\times d}$ by the multiplication of two matrices $W_1 \in \mathcal{R}^{d\times k}$ and $W_2 \in \mathcal{R}^{k\times d}$ with $k < \frac{d}{2}$. The rationale of this parametrization is that the mixing matrix tends to sum all the points in the point cloud, which makes its rank deficient. This point is fully validated in Figure~\ref{exp:scalability:clustering}. 

Following the trend of employing residual connection \cite{rn153} to boost gradient flow, we propose ResGraphX, which is a residual version of GraphX.
The main branch comprises of an FC layer (activated by ReLU) followed by a GraphX layer.
As in \cite{rn153}, the residual branch is an identity when the output dimension of the layer does not change, and an FC layer otherwise. When the upsampling version of ResGraphX, which shall be called UpResGraphX, is utilized, the residual branch has to be another GraphX to account for the expansion of the point set. In the deformation network, we employ three (Up)ResGraphX modules having 512, 256, and 128, respectively, and put a linear FC layer on top. Kindly refer to the supplementary for more technical details.

\section{Experimental results}
\textbf{Implementation details.}
We used Chamfer distance (CD) to measure the discrepancy between PCDNet's predictions and ground truths. For the sake of completeness, we write the CD between two point sets $\mathcal{X},\mathcal{Y}\subseteq \mathcal{R}^3$ below

\begin{equation}
\mathcal{L}(\mathcal{X},\mathcal{Y}) = \frac{1}{\left| \mathcal{X} \right|} \sum\limits_{x \in \mathcal{X}} {\mathop {\min }\limits_{y \in \mathcal{Y}} \left\| {x - y} \right\|_2^2}  + \frac{1}{\left| \mathcal{Y} \right|}\sum\limits_{y \in \mathcal{Y}} {\mathop {\min }\limits_{x \in \mathcal{X}} \left\| {y - x} \right\|_2^2} .
\end{equation}
The loss was optimized by the Adam optimizer \cite{rn65} with a learning rate of $5\text{e-}5$ and default exponential decay rates. To limit the function space, we incorporated a small ($1\text{e-}5$) L2 regularization term into the loss. We found that scheduling the learning rate helped to accelerate the optimization at the late stage, and so we multiplied it by $0.3$ at epochs 5 and 8. Training ran for totally 10 epochs in 3.5 days on a single NVIDIA TitanX 12GB RAM. Batch size of $4$ was used in all training scenarios.

At every iteration of the training, we initialized a random point cloud so that given fixed camera intrinsics, the projection of the point cloud covers the whole image plane. We used an initial point cloud of 2k points in all experiments if not otherwise specified.

\begin{figure*}[h!]
\centering
\includegraphics[width=\textwidth]{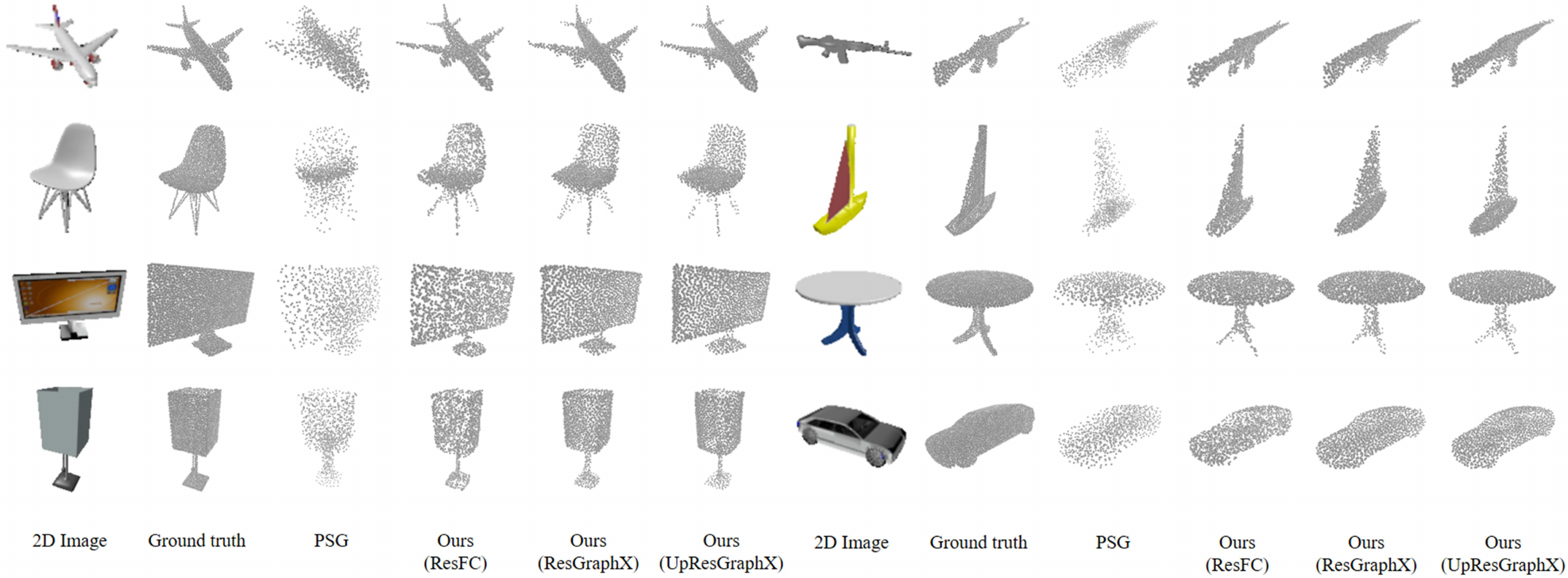}
\caption{Qualitative performance of PSG \cite{rn262} and different variants of PCDNet on ShapeNet. Our results are denser and more accurate than those produced by PSG.}
\label{exp:qualitative}
\end{figure*}

\begin{figure*}[h!]
\centering
\includegraphics[width=\textwidth]{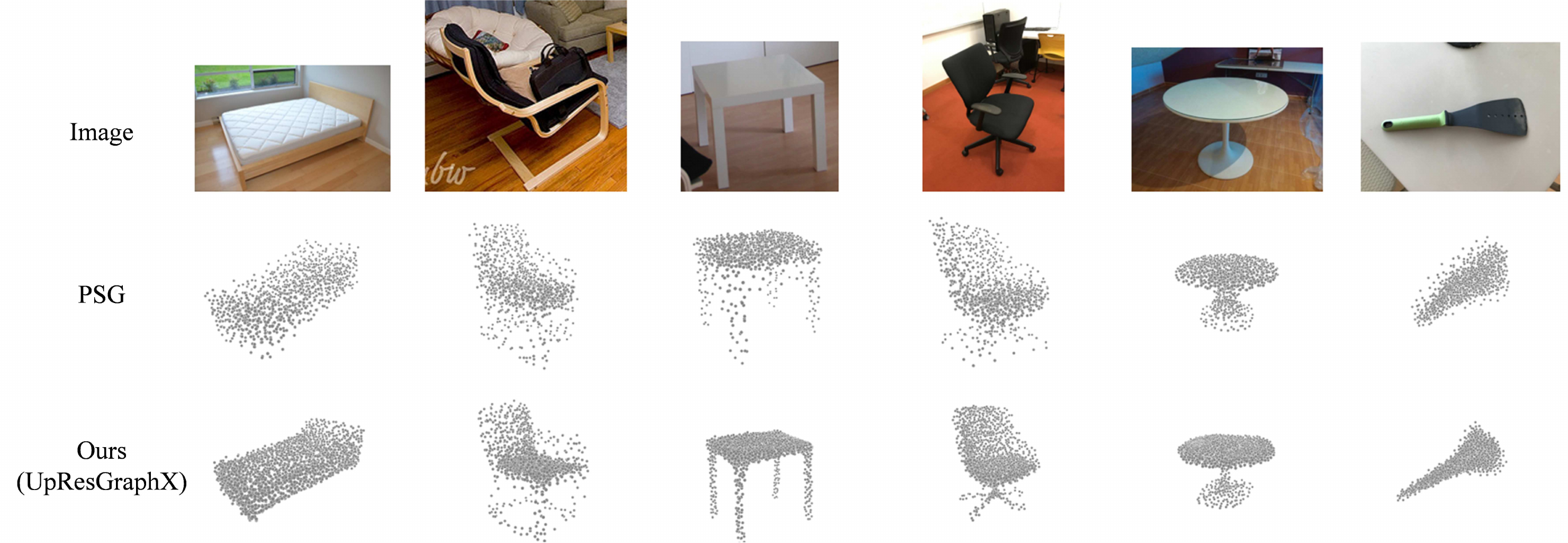}
\caption{Qualitative performance of PSG \cite{rn262} and PCDNet-UpResGraphX on some real life images taken from Pix3D. The predictions from PSG have high variance compared to ours, which present clear and solid shapes.}
\label{exp:qualitative-real}
\end{figure*}

\begin{table*}
\centering
\caption{Quantitative performance of different single image point cloud generation methods on 13 major categories of ShapeNet. ``$\uparrow$"~indicates higher is better. ``$\downarrow$" specifies the opposition. Best performance is highlighted in \textbf{bold}.}
\label{exp:quantitative}
\resizebox{\textwidth}{!}{%
\begin{tabular}{c|c|ccccccccccccc|c}
\hline
\multicolumn{2}{c|}{Category}                             & table          & car            & chair          & plane          & couch          & firearm        & lamp           & watercraft     & bench          & speaker        & cabinet        & monitor        & cellphone      & mean           \\ \hline
\multirow{8}{*}{CD$\downarrow$} & 3D-R2N2 \cite{rn272}    & 1.116          & 0.845          & 1.432          & 0.895          & 1.135          & 0.993          & 4.009          & 1.215          & 1.891          & 1.507          & 0.735          & 1.707          & 1.137          & 1.445          \\ \cline{2-16} 
                                & PSG \cite{rn262}        & 0.517          & 0.333          & 0.645          & 0.430          & 0.549          & 0.423          & 1.193          & 0.633          & 0.629          & 0.756          & 0.439          & 0.722          & 0.438          & 0.593          \\ \cline{2-16} 
                                & Pixel2mesh \cite{rn259} & 0.498          & 0.268          & 0.610          & 0.477          & 0.490          & 0.453          & 1.295          & 0.670          & 0.624          & 0.739          & 0.381          & 0.755          & 0.421          & 0.591          \\ \cline{2-16} 
                                & Ours (FC)               & 0.314          & 0.220          & 0.333          & 0.127          & 0.289          & 0.128          & 0.560          & 0.30           & 0.211          & 0.471          & 0.310          & 0.275          & 0.181          & 0.286          \\ \cline{2-16} 
                                & Ours (ResFC)            & 0.305          & 0.216          & 0.321          & 0.123          & 0.284          & 0.123          & 0.543          & 0.228          & 0.204          & 0.474          & 0.309          & 0.272          & 0.181          & 0.276          \\ \cline{2-16} 
                                & Ours (GraphX)           & 0.299          & 0.192          & 0.317          & 0.123          & 0.265          & 0.127          & 0.549          & 0.214          & 0.202          & 0.433          & 0.272          & 0.258          & 0.159          & 0.262          \\ \cline{2-16} 
                                & Ours (ResGraphX)        & 0.291          & 0.188          & 0.313          & 0.120          & 0.259          & 0.124          & 0.529          & 0.214          & 0.199          & 0.430          & 0.275          & 0.257          & 0.159          & 0.259          \\ \cline{2-16} 
                                & Ours (UpResGraphX)      & \textbf{0.284} & \textbf{0.184} & \textbf{0.306} & \textbf{0.116} & \textbf{0.254} & \textbf{0.119} & \textbf{0.523} & \textbf{0.210} & \textbf{0.189} & \textbf{0.419} & \textbf{0.265} & \textbf{0.248} & \textbf{0.155} & \textbf{0.252} \\ \hline
\multirow{8}{*}{IoU$\uparrow$}  & 3D-R2N2 \cite{rn272}    & 0.580          & \textbf{0.836} & 0.550          & 0.561          & 0.706          & 0.600          & 0.421          & 0.610          & 0.527          & 0.717          & 0.772          & 0.565          & 0.754          & 0.631          \\ \cline{2-16} 
                                & PSG \cite{rn262}        & 0.606          & 0.831          & 0.544          & 0.601          & 0.708          & 0.604          & 0.462          & 0.611          & 0.550          & \textbf{0.737} & 0.771          & 0.552          & 0.749          & 0.640          \\ \cline{2-16} 
                                & GAL \cite{rn278}        & \textbf{0.714} & 0.737          & 0.700          & 0.685          & 0.739          & 0.715          & \textbf{0.670} & 0.675          & 0.709          & 0.698          & 0.772          & \textbf{0.804} & 0.773          & 0.712          \\ \cline{2-16} 
                                & Ours (FC)               & 0.676          & 0.820          & 0.693          & 0.779          & 0.784          & 0.757          & 0.552          & 0.769          & 0.739          & 0.713          & 0.769          & 0.764          & 0.846          & 0.743          \\ \cline{2-16} 
                                & Ours (ResFC)            & 0.688          & 0.821          & \textbf{0.704} & \textbf{0.791} & 0.786          & \textbf{0.765} & 0.573          & \textbf{0.772} & \textbf{0.746} & 0.715          & 0.770          & 0.765          & 0.848          & \textbf{0.750} \\ \cline{2-16} 
                                & Ours (GraphX)           & 0.487          & 0.720          & 0.550          & 0.734          & 0.645          & 0.715          & 0.487          & 0.705          & 0.592          & 0.617          & 0.677          & 0.680          & 0.821          & 0.648          \\ \cline{2-16} 
                                & Ours (ResGraphX)        & 0.532          & 0.833          & 0.689          & 0.766          & \textbf{0.790} & 0.751          & 0.532          & 0.763          & 0.738          & 0.724          & \textbf{0.781} & 0.757          & \textbf{0.858} & 0.732          \\ \cline{2-16} 
                                & Ours (UpResGraphX)      & 0.605          & 0.819          & 0.663          & 0.758          & 0.770          & 0.747          & 0.516          & 0.754          & 0.725          & 0.708          & 0.770          & 0.735          & 0.857          & 0.725          \\ \hline
\end{tabular}%
}
\end{table*}

\textbf{Data.} We trained and evaluated our model on the ShapeNet dataset \cite{rn271}. ShapeNet is the largest collection of 3D CAD models that is publicly available. We used a subset of the ShapeNet core consisting of around 50k models categorized into 13 major groups. We utilized the default train/test split shipped with the database. All the hyperparameters were selected solely based on the convergence rate of the training loss. The rendered images and ground truth point clouds were kindly provided by \cite{rn272}. Different from previous works, we used only grayscale images as we found no clear benefit when using RGB. 

\textbf{Benchmarking methods.} We pitted our PCDNet against current state-of-the-art methods including 3D-R2N2 \cite{rn272}, point set generation network (PSG) \cite{rn262}, pixel-to-mesh (Pixel2mesh) \cite{rn259}, and geometric adversarial network (GAL) \cite{rn278}. 3D-R2N2 aimed to provide a unified framework for 3D reconstruction whether the problem is single-view or multi-view by harnessing a 3D RNN architecture. PSG is a regressor that directly converts an RGB image into point cloud, which is the most similar method of the four competing models. Pixel2mesh utilized the graph convolution to deform a predefined mesh into object shape given an RGB input. Finally, GAL resorted to adversarial loss \cite{rn92} and multi-view reprojection loss in addition to CD to estimate a representative point cloud.

\textbf{PCDNet variants.} We tested five variants of PCDNet: (1) a naive model with an FC deformation network, (2) a model with a residual FC (ResFC) deformation network, (3) a model with GraphX, (4) a model with ResGraphX, and (5) a model with UpResGraphX. For more details about the five architectures, see the supplementary and our website.

\textbf{Metrics.} To make it easier for PCDNet to serve as a baseline in subsequent research, we reported two common metric scores which are CD and intersection over union (IoU). CD is our main criterion, not because PCDNet is trained using CD, but it is better correlated with human perception \cite{rn280}. 
IoU quantifies the overlapping region between two input sets. Regarding IoU, we first voxelized the point sets into a $32\times 32\times 32$ grid and calculated the scores. We note that while PSG learns how to voxelize to achieve the best IoU and GAL is indirectly trained to maximize IoU, we used a simple voxelization method in \cite{rn263}.

\subsection{Comparison to state-of-the-art methods}

\subsubsection{Qualitative results}
We start by comparing the results obtained by PCDNet and PSG visually. The results are demonstrated in Figure~\ref{exp:qualitative}. As can be seen from the figure, even our naive formulation easily outperforms the competing method in all cases. While the estimated point clouds from PSG are very sparse and have high variance, those from PCDNet have pretty sharp and solid shapes. Our models preserve both the appearances and fine details much better thanks to the global and per-point features embedded in our proposed method.

We also tested our best model, PCDNet-UpResGraphX, on some real-world object images taken from Pix3D \cite{rn280}. We applied the provided masks to the object images and let the model predict the point cloud representations of the images. We also obtained the results from PSG by the same way\footnote{PSG provides a model taking the concatenation of image and mask as input but the results are actually worse.}. The scenario is challenging as the lighting and occlusion are far different from the CG images. Nevertheless, the results produced by PCDNet are surprisingly impressive. Obviously, our predictions are much more reliable as the shapes are precise and more recognizable than those from PSG. We highlight that the objects that are not chair or table are out-of-distribution as similar objects were not included in training. This suggests that our method is capable of analyzing and reasoning about shapes, and not just memorizing what it has seen during training.

\subsubsection{Quantitative results}

The metric scores of PCDNet versus others are tabulated in Table~\ref{exp:quantitative}. As anticipated, all PCDNet variants outrun all the competing methods by a huge gap. Specifically, the average CD scores from our simplest model (FC) is already twice better than the state of the art. For IoU, our method still tops the table and raises the performance bar which was previously set by GAL. 
Also, among all the variants of PCDNet, the GraphX family obtains better CD scores than the baseline whose deformation network is made of only FC layers. This is no surprise as GraphX is purposely architected to model both the global semantics and local relationship of points in the point cloud, which is necessary for characterizing point sets \cite{rn260,rn270}. 
On the other hand, a deformation network with (Res)FC layers treats every point almost independently (points are processed independently in the forward pass but gradients are collectively computed in the backward pass), so the output coordinates are predicted without conditioning on the semantic shape information nor local coherence, which certainly degrades the performance. 
Still and all, the gain in CD comes at the cost of lower IoU. This might suggest that to get the best of both worlds, a new loss function should be designed to simultaneously optimize the two metrics. A promising solution could be a combination of CD and a reprojection loss as in \cite{rn263} or \cite{rn278}. 

\begin{figure}
\centering
\includegraphics[width=0.5\textwidth]{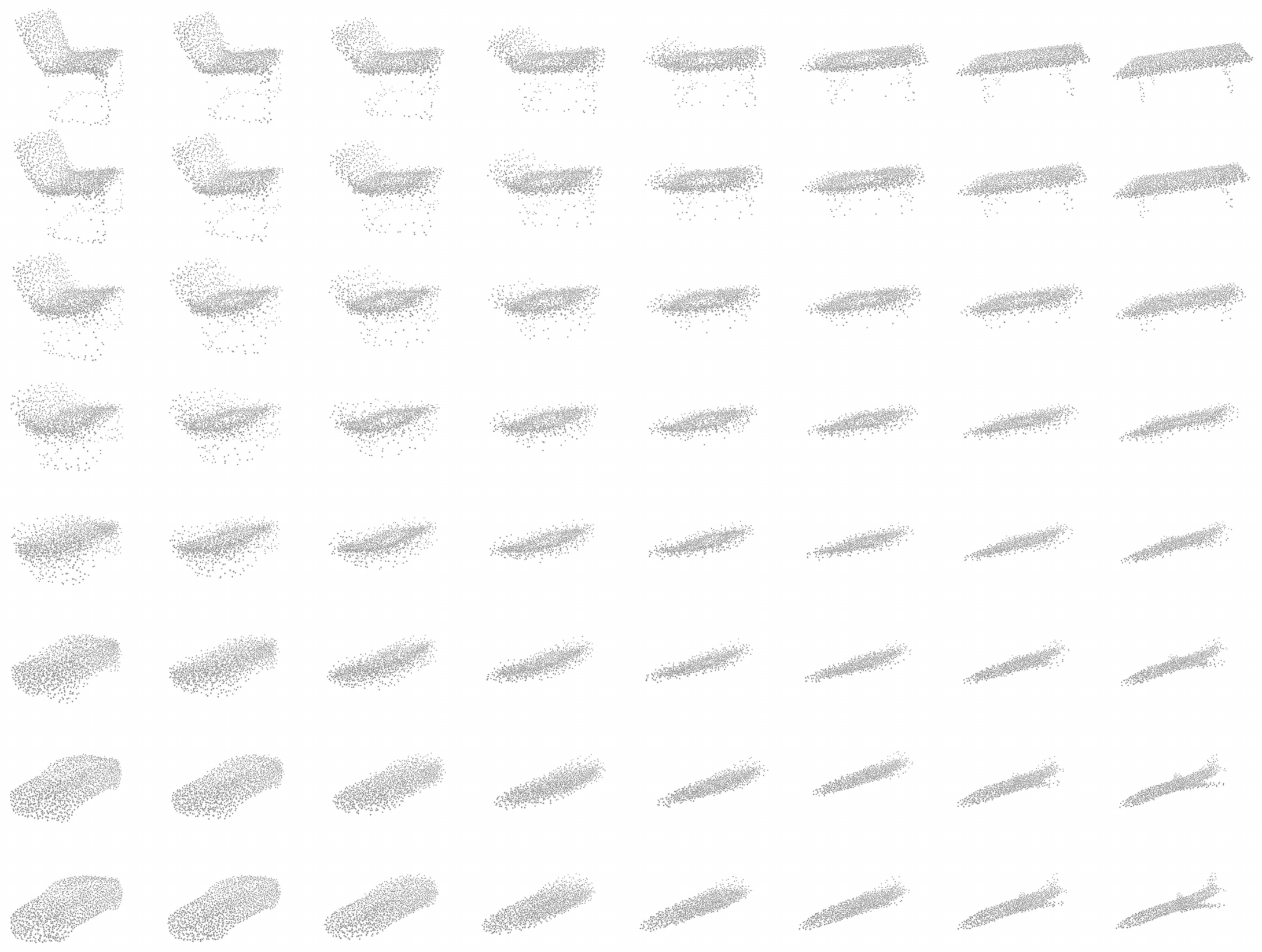}
\caption{Samples generated from interpolated latent representations of chair (top left), table (top right), car(bottom left), and airplane (bottom right).}
\label{exp:interpolation}
\end{figure}

To our surprise, the best performance is achieved by the model using UpResGraphX. This is intriguingly interesting because this model uses fewer parameters than other members in the GraphX family. 
We measured the multiply-accumulate (Mac) Flops for PCDNet-UpResGraphX\footnote{Using \url{https://git.io/fjHy9}.} and Pixel2mesh\footnote{Using tf.profile.}. Our model has only 1.91 GMac while Pixel2mesh has 1.95 GMac. From this result, the hypothesis that the performance is boosted thanks to the additional computing power can be ruled out.
We conjecture that other models slightly suffer from overfitting due to the large number of parameters. It is noted that the upsampling version of GraphX is potentially useful in point cloud upsampling, which covers the problem of point cloud densification. 

\subsection{Latent interpolation}

\begin{figure*}[h!]
\centering
\includegraphics[width=\textwidth]{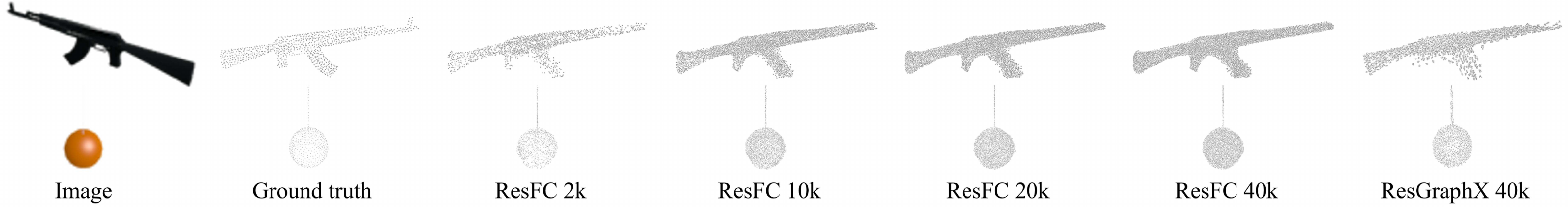}
\caption{Scalability of our method. Our model can produce arbitrarily dense point clouds by leveraging the stochasticity from the randomly generated point cloud in training.}
\label{exp:ablation:scalability}
\end{figure*}

\begin{figure}[!t]
   \centering
   \subfigure[]{\includegraphics[width=0.2\textwidth]{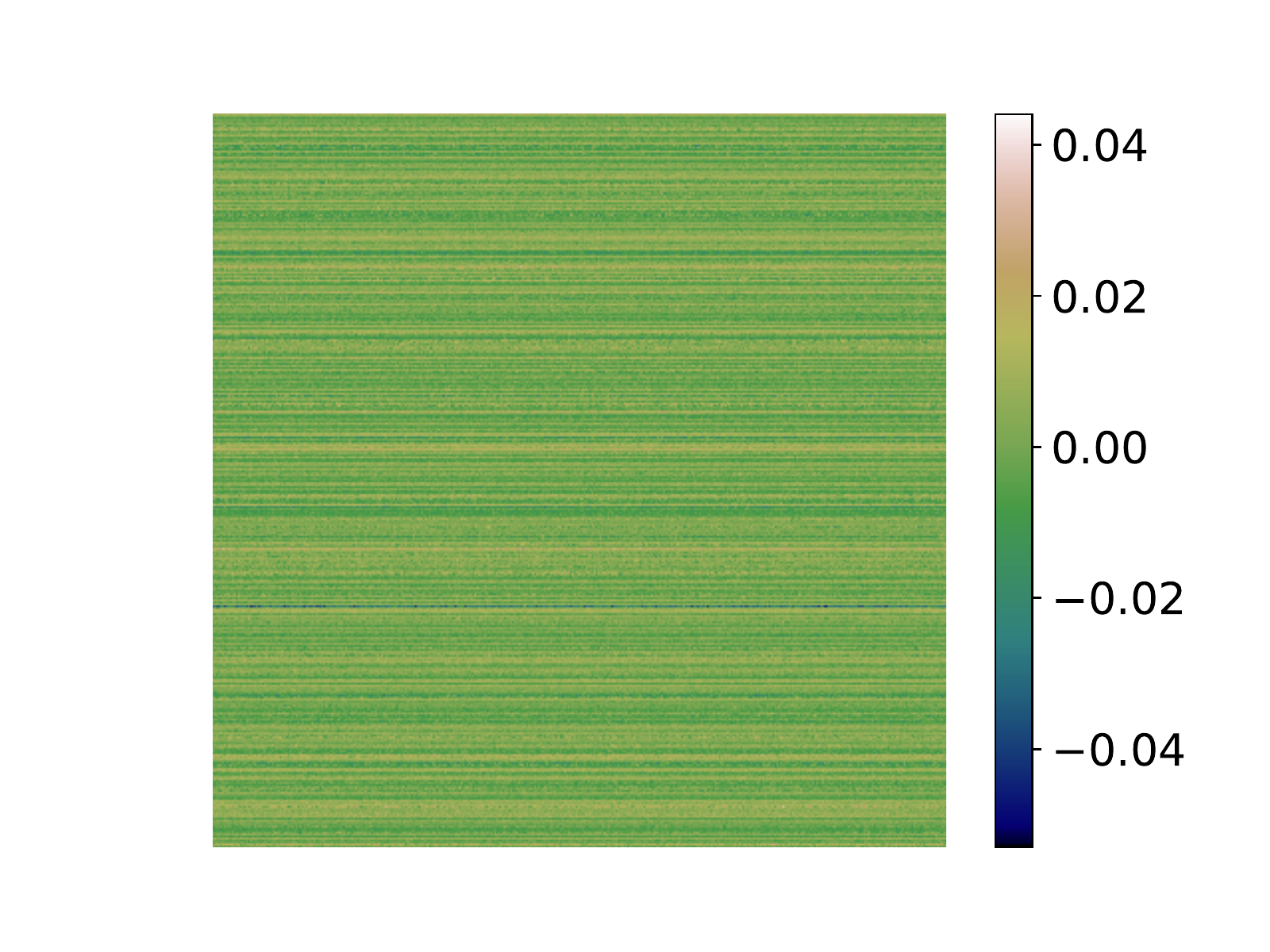}}
   \hfil
   \subfigure[]{\includegraphics[width=0.21\textwidth]{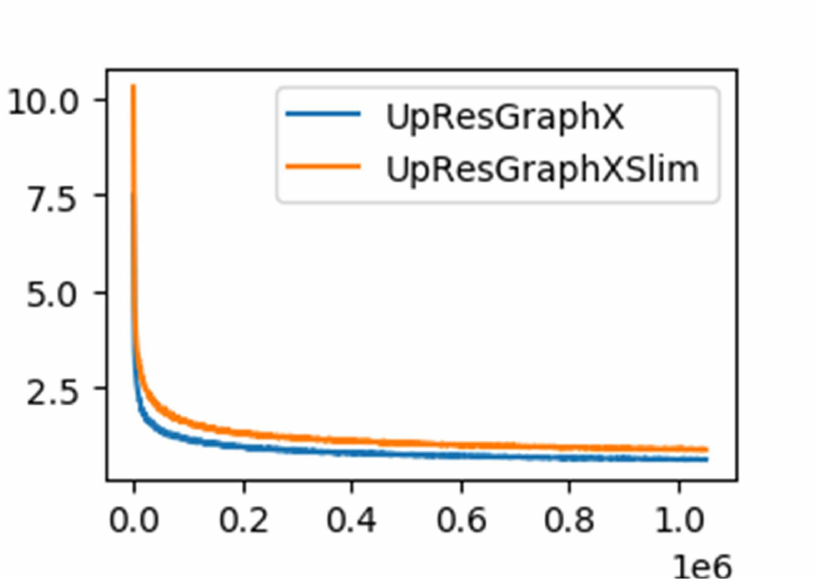}}
   \caption{(a) A mixing weight of a trained ResGraphX. (b) The training curves of PCDNet (UpResGraphX) and the simplified version (UpResGraphXSlim). Traning was terminated at epoch 5.}
   \label{exp:scalability:clustering}
\end{figure}

In this section, the latent representation from the feature extraction process is analyzed. 
We hypothesize that in order for the deformation network to generate an accurate point set representation, the latent must contain rich shape information. To illustrate our point, we conducted an interpolation experiment in latent space. We randomly chose four input images and obtained their latent representations following Section~\ref{method:feature}. Next, we synthesized a convex collection of 64 latent codes using bilinear interpolation. Finally, we decoded the codes and arranged the results in an $8\times 8$ grid as shown in Figure~\ref{exp:interpolation}. As can be seen, PCDNet smoothly interpolates between objects, either between similar objects such as chair and table or alien items like chair and airplane. Some semantics are also preserved; for \eg, chair legs are morphed into table legs. This proves that the network learns a smooth function and can generalize well over the object space, not simply putting mass on known objects.

\subsection{Scalability of PCDNet and analysis of GraphX}
To produce a dense point cloud, we batch several random point clouds and input it to PCDNet along with the input image. The outputs are then merged to obtain a unified point cloud. This is possible thanks to the stochasticity introduced by the random input cloud in each iteration during training. Figure~\ref{exp:ablation:scalability} shows off the scalability of PCDNet in which the point number ranges from 2k to 40k. As can be seen, the point cloud can be arbitrarily dense, unlike previous works which always have an upper bound for the set size. 

It can be noticed that the dense point clouds generated by the GraphX-series models cluster, which we shall refer to as \textit{clustering effect}. One example of such effect is shown in Figure~\ref{exp:ablation:scalability} (rightmost). To understand the problem, we plotted the mixing matrix of a trained model and observed an interesting phenomenon leading to the understanding of how GraphX probably works under the hood. The image is shown in Figure~\ref{exp:scalability:clustering}~(a). The barcode-like image suggests that apparently, GraphX lazily takes the mean of all feature vectors, and then learns to scale and shift it properly, which explains away the clustering effect. 
This also reinforces the choice of the sum operator used in \cite{Zaheer_Kottur_Ravanbakhsh_Poczos_Salakhutdinov_Smola_2017} to aggregate information. We conducted an experiment to verify this hypothesis and plotted the training curves in Figure~\ref{exp:scalability:clustering}~(b) (the hypothesized model is called UpResGraphXSlim.)
Even though the hypothesis seems to work, it learns much slower than the original version, and the gap between the two does not seem to be narrower as training advances. We admit that our hypothetical view of GraphX is just the tip of the iceberg, and there is still room for deeper theoretical analyses in subsequent work. 

\subsection{Ablation study}
\label{exp:ablation}

\textbf{Setup.} For time consideration, we experimented with only 5 major categories out of 13. Except for the ablated feature, all the options and hyperparameters are the same as in the main experiment.

\begin{figure}
\centering
\includegraphics[width=0.5\textwidth]{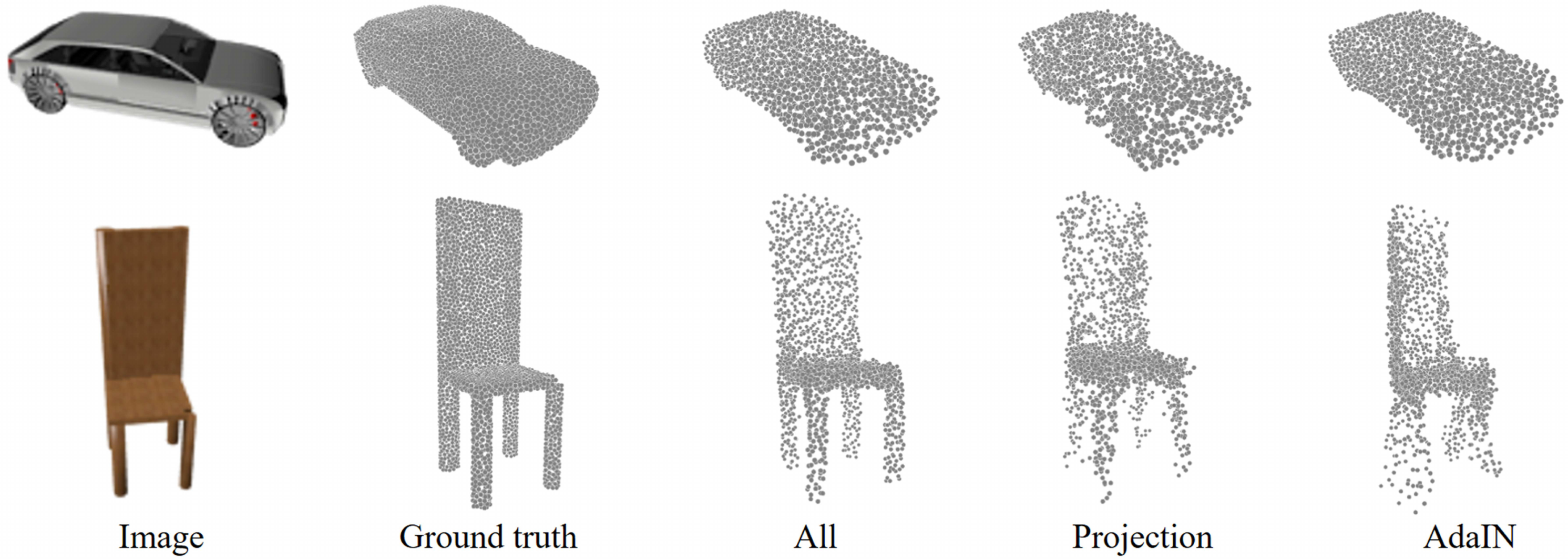}
\caption{Qualitative performance of PCDNet when different features are ablated from the model. This figure reveals exactly the contribution of each feature used in our formulation.}
\label{exp:ablation:fig}
\end{figure}

\textbf{Results.} Table~\ref{exp:ablation:table} and Figure~\ref{exp:ablation:fig} demonstrate the quantitative and qualitative results of the ablation study, respectively. As can be seen from the table, the models that incorporate only one feature (projection or AdaIN) achieves roughly the same performance. The projection feature helps PCDNet in CD score and AdaIN improves IoU. This can be easily explained by the fact that projection is a per-point feature that embraces the details of the shape, which is favored by a point-to-point metric like CD. On the other hand, IoU measures the coverage percentage of two volumetric models, which can be high when two objects have roughly the same shape but not necessary all the subtleties. When the two are combined, both the two scores are significantly boosted, which validates our design of PCDNet. 

The visualization in Figure~\ref{exp:ablation:fig} clearly illustrates the effect of each feature on the outputs. While the AdaIN feature helps to correctly model the global shapes (recognizable car and chair shapes), it lacks the information that can provide fine details. This is exactly opposite to the projection feature when it can precisely estimate some intricate model parts (for \eg, chair legs) but the overall appearances are not as solid as those from AdaIN. The combination of the two features provides a balance between fine details and global view, which enables our method to top every benchmark.

\begin{table}[]
\centering
\caption{Quantitative performance of PCDNet when different features are ablated.}
\label{exp:ablation:table}
\resizebox{0.47\textwidth}{!}{%
\begin{tabular}{c|c|ccccc|c}
\hline
\multicolumn{2}{c|}{Category} & table & car & chair & plane & lamp & mean \\ \hline
\multirow{3}{*}{CD$\downarrow$} & Ours (projection) & 0.637 & 0.284 & 0.490 & 0.177 & 0.670 & 0.452 \\ \cline{2-8} 
 & Ours (AdaIN) & \multicolumn{1}{l}{0.372} & \multicolumn{1}{l}{0.222} & \multicolumn{1}{l}{0.703} & \multicolumn{1}{l}{0.243} & \multicolumn{1}{l|}{0.564} & \multicolumn{1}{l}{0.421} \\ \cline{2-8} 
 & Ours (full) & 0.301 & 0.195 & 0.319 & 0.124 & 0.550 & 0.298 \\ \hline
\multirow{3}{*}{IoU$\uparrow$} & Ours (projection) & 0.540 & 0.818 & 0.657 & 0.704 & 0.501 & \multicolumn{1}{l}{0.644} \\ \cline{2-8} 
 & Ours (AdaIN) & \multicolumn{1}{l}{0.651} & \multicolumn{1}{l}{0.840} & \multicolumn{1}{l}{0.575} & \multicolumn{1}{l}{0.667} & \multicolumn{1}{l|}{0.523} & \multicolumn{1}{l}{0.651} \\ \cline{2-8} 
 & Ours (full) & 0.694 & 0.844 & 0.725 & 0.750 & 0.566 & 0.716 \\ \hline
\end{tabular}%
}
\end{table}

\section{Conclusion}
In this paper, we presented PCDNet, an architecture that deforms a random point set according to an input object image and produces a point cloud of the object. To deform the random point cloud, we first extracted global and per-point features for every point. While the point-specific features were obtained via projection following a previous work, the global features were distilled by AdaIN, a concept borrowed from style transfer literature. Having the features for each point, we deformed the point cloud by a network consisting of GraphX, a new layer that took into account the inter-correlation between points. The experiments validated the efficacy of the proposed method, which set a new height for single image 3D reconstruction. 

\section*{Acknowledgement}
This work was supported by Samsung Research Funding Center of Samsung Electronics under Project Number SRFC-IT1702-08.

{\small
\bibliographystyle{ieee_fullname}
\bibliography{graphx-conv}

\begin{thebibliography}{10}\itemsep=-1pt

\bibitem{rn273}
John Aloimonos.
\newblock Shape from texture.
\newblock {\em Biological cybernetics}, 58(5):345--360, 1988.

\bibitem{rn271}
Angel~X Chang, Thomas Funkhouser, Leonidas Guibas, Pat Hanrahan, Qixing Huang,
  Zimo Li, Silvio Savarese, Manolis Savva, Shuran Song, and Hao Su.
\newblock {S}hape{N}et: An information-rich 3{D} model repository.
\newblock {\em arXiv preprint arXiv:1512.03012}, 2015.

\bibitem{rn272}
Christopher~B Choy, Danfei Xu, JunYoung Gwak, Kevin Chen, and Silvio Savarese.
\newblock 3{D}-{R}2{N}2: A unified approach for single and multi-view 3{D}
  object reconstruction.
\newblock In {\em Proceedings of the {E}uropean {C}onference on {C}omputer
  {V}ision ({E}{C}{C}{V})}, pages 628--644. Springer, 2016.

\bibitem{rn262}
Haoqiang Fan, Hao Su, and Leonidas~J Guibas.
\newblock A point set generation network for 3{D} object reconstruction from a
  single image.
\newblock In {\em Proceedings of the IEEE conference on {C}omputer {V}ision and
  {P}attern {R}ecognition ({C}{V}{P}{R})}, pages 605--613, 2017.

\bibitem{rn92}
Ian Goodfellow, Jean Pouget-Abadie, Mehdi Mirza, Bing Xu, David Warde-Farley,
  Sherjil Ozair, Aaron Courville, and Yoshua Bengio.
\newblock Generative adversarial nets.
\newblock In {\em Advances in {N}eural {I}nformation {P}rocessing {S}ystems
  ({N}eur{I}{P}{S})}, pages 2672--2680, 2014.

\bibitem{rn153}
Kaiming He, Xiangyu Zhang, Shaoqing Ren, and Jian Sun.
\newblock Deep residual learning for image recognition.
\newblock In {\em Proceedings of the IEEE conference on {C}omputer {V}ision and
  {P}attern {R}ecognition ({C}{V}{P}{R})}, pages 770--778, 2016.

\bibitem{rn274}
Derek Hoiem, Alexei~A Efros, and Martial Hebert.
\newblock Automatic photo pop-up.
\newblock {\em ACM Transactions on Graphics}, 24(3):577--584, 2005.

\bibitem{rn218}
Xun Huang and Serge~J Belongie.
\newblock Arbitrary style transfer in real-time with adaptive instance
  normalization.
\newblock In {\em Proceedings of the {I}nternational {C}onference on {C}omputer
  {V}ision ({I}{C}{C}{V})}, pages 1510--1519, 2017.

\bibitem{rn263}
Eldar Insafutdinov and Alexey Dosovitskiy.
\newblock Unsupervised learning of shape and pose with differentiable point
  clouds.
\newblock In {\em Advances in {N}eural {I}nformation {P}rocessing {S}ystems
  ({N}eur{I}{P}{S})}, pages 2807--2817, 2018.

\bibitem{rn278}
Li Jiang, Shaoshuai Shi, Xiaojuan Qi, and Jiaya Jia.
\newblock {G}{A}{L}: Geometric adversarial loss for single-view 3{D}-object
  reconstruction.
\newblock In {\em Proceedings of the {E}uropean {C}onference on {C}omputer
  {V}ision ({E}{C}{C}{V})}, pages 802--816, 2018.

\bibitem{Kim_Lee_2017}
Jongyoo Kim and Sanghoon Lee.
\newblock Fully deep blind image quality predictor.
\newblock {\em IEEE Journal of Selected Topics in Signal Processing},
  11(1):206–220, 2017.

\bibitem{Kim_Nguyen_Ahn_Luo_Lee_2018}
Jongyoo Kim, Anh-Duc Nguyen, Sewoong Ahn, Chong Luo, and Sanghoon Lee.
\newblock Multiple level feature-based universal blind image quality assessment
  model.
\newblock In {\em 2018 25th IEEE International Conference on Image Processing
  (ICIP)}, page 291–295. IEEE, 2018.

\bibitem{Kim_Nguyen_Lee_2018}
Jongyoo Kim, Anh-Duc Nguyen, and Sanghoon Lee.
\newblock Deep cnn-based blind image quality predictor.
\newblock {\em IEEE Transactions on Neural Networks and Learning Systems},
  PP(99):1–14, 2018.

\bibitem{Kim_Zeng_Ghadiyaram_Lee_Zhang_Bovik_2017}
Jongyoo Kim, Hui Zeng, Deepti Ghadiyaram, Sanghoon Lee, Lei Zhang, and Alan~C
  Bovik.
\newblock Deep convolutional neural models for picture-quality prediction:
  Challenges and solutions to data-driven image quality assessment.
\newblock {\em IEEE Signal Processing Magazine}, 34(6):130–141, 2017.

\bibitem{Kim_Kim_Ahn_Kim_Lee_2018}
Woojae Kim, Jongyoo Kim, Sewoong Ahn, Jinwoo Kim, and Sanghoon Lee.
\newblock Deep video quality assessor: From spatio-temporal visual sensitivity
  to a convolutional neural aggregation network.
\newblock In {\em Proceedings of the European Conference on Computer Vision
  (ECCV)}, page 219–234, 2018.

\bibitem{rn65}
Diederik~P. Kingma and Jimmy Ba.
\newblock Adam: A method for stochastic optimization.
\newblock In {\em Proceedings of the International Conference on Learning
  Representations (ICLR)}, 2014.

\bibitem{rn261}
Thomas~N Kipf and Max Welling.
\newblock Semi-supervised classification with graph convolutional networks.
\newblock {\em arXiv preprint arXiv:1609.02907}, 2016.

\bibitem{rn260}
Yangyan Li, Rui Bu, Mingchao Sun, Wei Wu, Xinhan Di, and Baoquan Chen.
\newblock Pointcnn: Convolution on x-transformed points.
\newblock In {\em Advances in {N}eural {I}nformation {P}rocessing {S}ystems
  ({N}eur{I}{P}{S})}, pages 828--838, 2018.

\bibitem{rn276}
Chen-Hsuan Lin, Chen Kong, and Simon Lucey.
\newblock Learning efficient point cloud generation for dense 3{D} object
  reconstruction.
\newblock In {\em AAAI Conference on Artificial Intelligence}, 2018.

\bibitem{rn237}
Alexander Mordvintsev, Nicola Pezzotti, Ludwig Schubert, and Chris Olah.
\newblock Differentiable image parameterizations.
\newblock {\em Distill}, 3(7):e12, 2018.

\bibitem{Nguyen_Choi_Kim_Lee_2019}
Anh-Duc Nguyen, S Choi, W Kim, and S Lee.
\newblock A simple way of multimodal and arbitrary style transfer.
\newblock In {\em ICASSP 2019 - 2019 IEEE International Conference on
  Acoustics, Speech and Signal Processing (ICASSP)}, page 1752–1756, 2019.

\bibitem{rn265}
Emmanuel Prados and Olivier Faugeras.
\newblock {\em Shape from shading}, pages 375--388.
\newblock Springer, 2006.

\bibitem{rn270}
Charles~R Qi, Hao Su, Kaichun Mo, and Leonidas~J Guibas.
\newblock Pointnet: Deep learning on point sets for 3{D} classification and
  segmentation.
\newblock In {\em Proceedings of the IEEE Conference on {C}omputer {V}ision and
  {P}attern {R}ecognition ({C}{V}{P}{R})}, pages 652--660, 2017.

\bibitem{rn266}
Ashutosh Saxena, Min Sun, and Andrew~Y Ng.
\newblock Make3{D}: Learning 3{D} scene structure from a single still image.
\newblock {\em IEEE transactions on {P}attern {A}nalysis and {M}achine
  {I}ntelligence}, 31(5):824--840, 2009.

\bibitem{rn264}
Johannes~L Schonberger and Jan-Michael Frahm.
\newblock Structure-from-motion revisited.
\newblock In {\em Proceedings of the IEEE conference on {C}omputer {V}ision and
  {P}attern {R}ecognition ({C}{V}{P}{R})}, pages 4104--4113, 2016.

\bibitem{rn64}
Karen Simonyan and Andrew Zisserman.
\newblock Very deep convolutional networks for large-scale image recognition.
\newblock {\em arXiv preprint arXiv:1409.1556}, 2014.

\bibitem{rn280}
Xingyuan Sun, Jiajun Wu, Xiuming Zhang, Zhoutong Zhang, Chengkai Zhang, Tianfan
  Xue, Joshua~B Tenenbaum, and William~T Freeman.
\newblock Pix3{D}: Dataset and methods for single-image 3{D} shape modeling.
\newblock In {\em Proceedings of the IEEE Conference on {C}omputer {V}ision and
  {P}attern {R}ecognition ({C}{V}{P}{R})}, pages 2974--2983, 2018.

\bibitem{rn277}
Shubham Tulsiani, Alexei~A Efros, and Jitendra Malik.
\newblock Multi-view consistency as supervisory signal for learning shape and
  pose prediction.
\newblock In {\em Proceedings of the IEEE Conference on {C}omputer {V}ision and
  {P}attern {R}ecognition ({C}{V}{P}{R})}, pages 2897--2905, 2018.

\bibitem{rn275}
Shubham Tulsiani, Tinghui Zhou, Alexei~A Efros, and Jitendra Malik.
\newblock Multi-view supervision for single-view reconstruction via
  differentiable ray consistency.
\newblock In {\em Proceedings of the IEEE conference on {C}omputer {V}ision and
  {P}attern {R}ecognition ({C}{V}{P}{R})}, pages 2626--2634, 2017.

\bibitem{rn259}
Nanyang Wang, Yinda Zhang, Zhuwen Li, Yanwei Fu, Wei Liu, and Yu-Gang Jiang.
\newblock Pixel2mesh: Generating 3{D} mesh models from single rgb images.
\newblock In {\em Proceedings of the {E}uropean {C}onference on {C}omputer
  {V}ision ({E}{C}{C}{V})}, pages 52--67, 2018.

\bibitem{rn269}
Jiajun Wu, Chengkai Zhang, Tianfan Xue, Bill Freeman, and Josh Tenenbaum.
\newblock Learning a probabilistic latent space of object shapes via 3{D}
  generative-adversarial modeling.
\newblock In {\em Advances in {N}eural {I}nformation {P}rocessing {S}ystems
  ({N}eur{I}{P}{S})}, pages 82--90, 2016.

\bibitem{rn267}
Zhirong Wu, Shuran Song, Aditya Khosla, Fisher Yu, Linguang Zhang, Xiaoou Tang,
  and Jianxiong Xiao.
\newblock 3{D} {S}hape{N}ets: A deep representation for volumetric shapes.
\newblock In {\em Proceedings of the IEEE conference on {C}omputer {V}ision and
  {P}attern {R}ecognition ({C}{V}{P}{R})}, pages 1912--1920, 2015.

\bibitem{rn268}
Xinchen Yan, Jimei Yang, Ersin Yumer, Yijie Guo, and Honglak Lee.
\newblock Perspective transformer nets: Learning single-view 3{D} object
  reconstruction without 3{D} supervision.
\newblock In {\em Advances in {N}eural {I}nformation {P}rocessing {S}ystems
  ({N}eur{I}{P}{S})}, pages 1696--1704, 2016.

\bibitem{Zaheer_Kottur_Ravanbakhsh_Poczos_Salakhutdinov_Smola_2017}
Manzil Zaheer, Satwik Kottur, Siamak Ravanbakhsh, Barnabas Poczos, Ruslan~R
  Salakhutdinov, and Alexander~J Smola.
\newblock Deep sets.
\newblock In {\em Advances in Neural Information Processing Systems (NeurIPS)},
  page 3391–3401, 2017.

\end{thebibliography}
}

\end{document}